\documentclass{article}
\usepackage{spconf,amsmath,graphicx,hyperref}
\usepackage{amssymb} 
\usepackage{enumitem}
\usepackage{booktabs}
\usepackage{multirow}
\usepackage{graphicx} 
\usepackage{lipsum} 
\usepackage{xcolor} 

\title{DP-DeGauss: Dynamic Probabilistic Gaussian Decomposition for Egocentric 4D Scene Reconstruction}
\name{Tingxi Chen$^{1}$\quad Zhengxue Cheng$^{1}$ \quad Houqiang Zhong$^{1}$\quad Su Wang$^{2}$\quad Rong Xie$^{1}$\quad Li Song$^{1}$}
\address{$^{1}$Institute of Image Communication and Network Engineering, Shanghai Jiao Tong University \\ $^{2}$Nvidia.\\{\tt\small \{ctx.ximi17,zxcheng,zhonghouqiang,xierong,song\_li\}@sjtu.edu.cn},{\tt\small suwang@Nvidia.com }}

\begin{document}
\ninept
\maketitle
\begin{abstract}
Egocentric video is crucial for next-generation 4D scene reconstruction, with applications in AR/VR and embodied AI. However, reconstructing dynamic first-person scenes is challenging due to complex ego-motion, occlusions, and  hand–object interactions. Existing decomposition methods are ill-suited, assuming fixed viewpoints or merging dynamics into a single foreground. To address these limitations, we introduce DP-DeGauss, a dynamic probabilistic Gaussian decomposition framework for egocentric 4D reconstruction. Our method initializes a unified 3D Gaussian set from COLMAP priors, augments each  with a learnable category probability, and dynamically routes them into specialized deformation branches for background, hands, or object modeling. We employ category-specific masks for better disentanglement and introduce brightness and motion-flow control to improve static rendering and dynamic reconstruction.Extensive experiments show that DP-DeGauss outperforms baselines by +1.70dB in PSNR on average with  SSIM and LPIPS gains. More importantly, our framework achieves the first and state-of-the-art disentanglement of background, hand, and object components, enabling explicit, fine-grained separation, paving the way for more intuitive ego scene understanding and editing.

\end{abstract}

\section{Introduction}
\label{sec:intro}

Egocentric video offers a unique window into human activities, capturing continuous interactions between hands, objects, and the surrounding environment. With the rise of large-scale egocentric datasets, researchers have begun exploring 4D reconstruction and interaction modeling from this perspective \cite{hoi4d,epic,hot3d,aea,adt}. However, dynamic scene reconstruction from egocentric videos remains highly challenging. Unlike static multi-view captures, egocentric data features strong camera motion, frequent occlusions, and complex hand–object interactions. These factors hinder clean reconstruction, let alone fine-grained disentanglement of hands, manipulated objects, and static backgrounds.

Recent advances in Neural Radiance Fields \cite{nerf} and 3D Gaussian Splatting \cite{3dgs} have enabled scalable dynamic reconstruction. We adopt 3DGS for its high-quality and efficient rasterization. While originally designed for static scenes, dynamic extensions \cite{4dgs,deformablegs,sun2025sdd} introduce deformation networks or HexPlane encoders with MLP decoders to model time-varying Gaussian deformations. However, these approaches treat the entire scene with a single network, increasing computational cost and preventing independent motion learning. \cite{swift4d} improves efficiency via static–dynamic decomposition, but its pixel-intensity masks assume fixed viewpoints and fail on egocentric videos. \cite{egogaussian} manually splits videos into static and dynamic clips, which is labor-intensive and slow. \cite{degauss} automatically separates dynamic and static components using depth cues, but initializes dynamic regions with random Gaussians that underutilize point cloud priors and only achieve coarse foreground–background separation, leaving hand–object separation unresolved. Overall, existing methods either assume fixed viewpoints, require manual annotations, or restrict decomposition only to static v.s. dynamic levels—insufficient for egocentric scenarios where robust initialization and fine-grained hand–object-background disentanglement are essential.
\begin{figure}[t]
    \centering
    \includegraphics[width=1\linewidth]{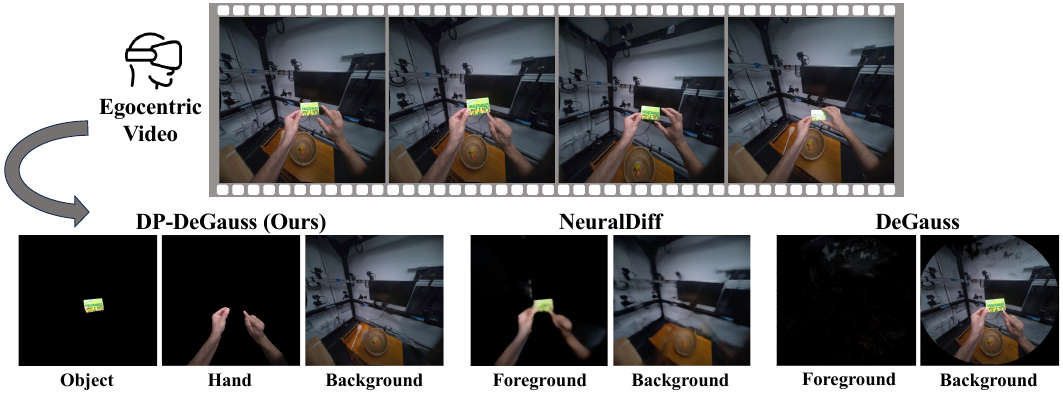}
    \caption{Fine-grained decomposition of egocentric scenes with DP-DeGauss. We achieve accurate and clean separation of background, hands, and objects, overcoming prior methods’ limitations of low-detail reconstruction, misclassification, and coarse foreground–background separation without hand–object distinction.
}
    \label{fig:teaser}
\end{figure}
\begin{figure*}[t]
    \centering
    \includegraphics[width=1\linewidth]{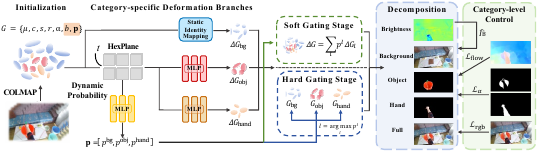}
    \caption{Overview of our proposed \textbf{DP-DeGauss}. A unified Gaussian set initialized from COLMAP is augmented with a learnable brightness attribute and dynamic category probabilities, which route points to category-specific deformation branches via a two-stage soft-to-hard gating process. Category-level controls jointly drive accurate reconstruction and fine-grained decomposition.}
    \label{fig:pipeline}
\end{figure*}

Meanwhile, existing hand–object reconstruction methods rely on predefined 3D models, sophisticated optimization, or carefully calibrated multi-view setups \cite{handdiffusion,zhu2023get,yang2021cpf,hold,novel}, limiting scalability in unconstrained egocentric settings with large motion and occlusions.

To this end, we propose DP-DeGauss, dynamic probabilistic Gaussian decomposition framework with a soft-to-hard strategy for egocentric reconstruction without hand or object priors. We leverage Structure From Motion (SFM) priors for an unified scene Gaussian set \cite{colmap}, augmenting each with a learnable probabilistic 
 and brightness attribute. A lightweight MLP dynamically estimates class probabilities  (background, hand, or object), enabling robust initialization and adaptive decomposition into category-specific deformation branches. We further incorporate segmentation masks for supervision, brightness control  for stable background rendering, and optical flow constraints to refine hand/object dynamics \cite{degauss,motiongs}. This jointly achieves holistic 4D reconstruction and explicit instance-level separation as in Fig.\ref{fig:teaser}, advancing egocentric scene understanding.

Our main contributions can be summarized as follows:
\begin{itemize}[leftmargin=10pt, labelsep=0.5em, itemsep=0pt, topsep=0pt, parsep=0pt]
\item We introduce \textbf{DP-DeGauss}, a dynamic probabilistic Gaussian decomposition framework, resolving egocentric initialization difficulties while enabling a soft-to-hard adaptive decomposition into background, hand, and object branches.
\item We introduce category-specific controls: brightness regulation for background, motion-flow guidance for dynamic hands/objects, and segmentation mask supervision for instance boundaries, enhancing static stability and dynamic fidelity in reconstruction.
\item We deliver high-quality reconstruction with fine-grained disentanglement of background and interacting components in egocentric scenarios.
\end{itemize}

\section{Methods}

\label{sec:Methods}
Our method (Fig.\ref{fig:pipeline}) is a dynamic probabilistic Gaussian decomposition framework from soft to hard for egocentric 4D scene reconstruction. Starting from the standard 3D Gaussian Splatting, we propose a unified Gaussian representation with learnable category probabilities for background, hand, and object, followed by category-level control strategies to enhance reconstruction quality and separation.
 \subsection{Preliminary: 3D Gaussian Splatting}
Every 3D Gaussian is defined by its center $\mu \in \mathbb{R}^3$, covariance $\Sigma\in \mathbb{R}^{3 \times 3}$ (parameterized by rotation $r$ and scaling $s$, ), color $c\in \mathbb{R}^k$ ( \textit{k }is numbers of SH functions), and opacity $\alpha \in \mathbb{R}$. 
The spatial density is:
\begin{equation}G(\mathbf{x})=e^{-\frac{1}{2}(\mathbf{x}-\mu)^T\Sigma^{-1}(\mathbf{x}-\mu)}\end{equation}
Depth‑ordered Gaussians  $\mathcal{N}$ are composited for differentiable rendering using the front‑to‑back rule:

\begin{equation}
I = \sum_{i \in \mathcal{N}} \alpha_i \, c_i\prod_{j < i} \big( 1 - \alpha_j \big),
\end{equation}

\subsection{Dynamic Probabilistic Gaussian Decomposition}
\label{sec:Probabilistic}
\begin{figure*}[thbp]
    \centering
    \includegraphics[width=1\linewidth]{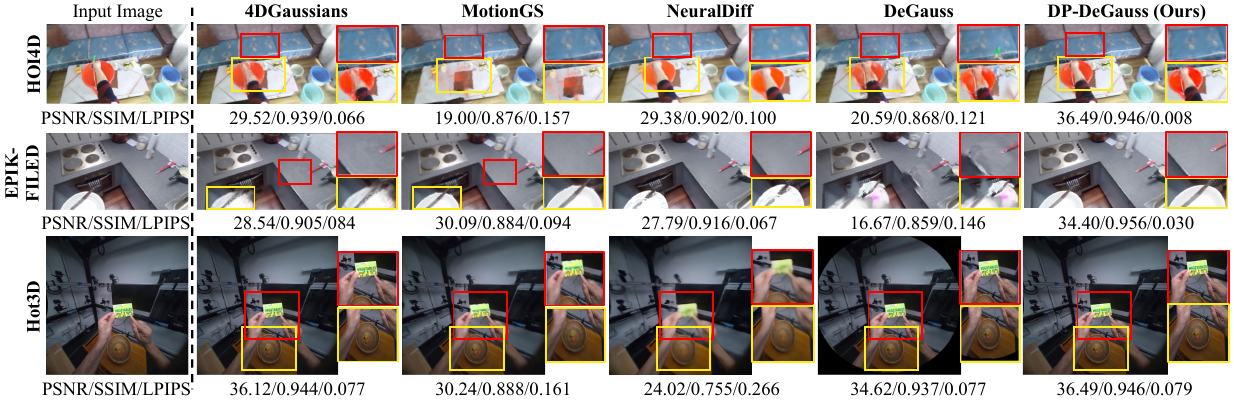}
    \caption{Qualitative comparison of full-scene reconstruction. Our approach yields sharper geometry, reduced motion blur, and fewer holes in both static background and dynamic object/hand regions.}
    \label{fig:Reconstruction}
\end{figure*}

In contrast to approaches that pre‑segment static and dynamic components or initialize them separately and randomly—thus failing to fully exploit priors—we initialize a unified Gaussian cloud from SFM covering background, hands, and objects without separation. To jointly encode static and dynamic components in this unified set, we extend each standard Gaussian point \cite{3dgs} to:
\begin{equation}
    G=\left\{\mu, c, s, r, \alpha, b, \mathbf{p}\right\}
\end{equation}
where each Gaussian is augmented with a brightness control attribute $b$ (introduced in \ref{sec:control}) and a dynamic probability vector $\mathbf{p}$ guiding subsequent deformation and decomposition:
\begin{equation}
    \mathbf{p}=\left[p^{\mathrm{bg}}, p^{\mathrm{obj}}, p^{\mathrm{hand}}\right], \quad \sum_{l \in\{\mathrm{bg},\mathrm{obj},\mathrm{hand}\}} p^{l}=1
\end{equation}
This vector encodes the likelihood that Gaussian $i$ belongs to background, object, or hand. At initialization, we assign a high prior to background, e.g., $\mathbf{p}=[0.8,0.1,0.1]$.

To capture motion in dynamic components while keeping the background static, we employ three category-specific deformation branches $\mathcal{F}_{\text{bg}}, \mathcal{F}_{\text{obj}}, \mathcal{F}_{\text{hand}}$ to compute time-aware Gaussian deformations $\Delta G_l = \mathcal{F}_l(G_l,t)$, and a global shared probability branch $\mathcal{P}$ to update $\mathbf{p}$ via $\Delta \mathbf{p}$
. The background branch $\mathcal{F}_{\text{bg}}$ is implemented as an identity mapping outputting no deformation, preserving static structure. The object and hand branches $\mathcal{F}_{\text{obj}}, \mathcal{F}_{\text{hand}}$ share a spatio-temporal HexPlane encoder $\mathcal{H}$ with parameters $\boldsymbol{\theta} = \{\mu, c, s, r, \alpha, b\}$, but use different MLP decoders $\mathcal{D}_l$ to predict $\Delta G_l = \mathcal{D}_l(\mathcal{H}(\boldsymbol{\theta}, t))$. The shared classification head $\mathcal{P}(\cdot)$, also an MLP, outputs $\Delta\mathbf{p} = \mathcal{P}(\mathcal{H}(\boldsymbol{\theta}, t))$ to update the category probability $\mathbf{p} \leftarrow \mathbf{p} + \Delta\mathbf{p}$.

We train in \textbf{two stages}. In the \textit{soft gating} stage, all deformation branches process all Gaussians, and $\mathbf{p}$ modulates the contributions of each branch:
\begin{equation}
    \Delta G = p^{\mathrm{bg}}\cdot\Delta G_{\mathrm{bg}} + p^{\mathrm{obj}}\cdot\Delta G_{\mathrm{obj}} + p^{\mathrm{hand}}\cdot\Delta G_{\mathrm{hand}}
\end{equation}
updating $G' = G + \Delta G$. This ensures every branch receives gradients even when $\mathbf{p}$ is still inaccurate, enabling continuous refinement of category assignments under photometric and mask supervision.

In the \textit{hard gating} stage, each Gaussian is exclusively assigned to its most probable category:
\begin{equation}
    \hat{l} = \underset{l}{\arg\max} \; p^l
\end{equation}
and routed exclusively through $\mathcal{F}_{\hat{l}}$ for deformation, preventing cross-branch influence. Unlike the soft stage, Gaussians here contribute solely to one category's attributes, but the global classification head $\mathcal{P}(\cdot)$ remains active to correct misclassifications via supervision.



 
For rendering, we produce both a composite image of full categories and separate per-category images. In the soft stage, where category assignments are given by soft probabilities $p^l$, all Gaussians contribute to each category rendering, with their opacity scaled by the corresponding probability. 
The per-category image for category $l$ is:
\begin{equation}
I_l = \sum_{i \in \mathcal{N}} ( p_i^{l} \,\alpha_i' )c_i' \prod_{j<i} (1 - p_j^{l} \alpha_j' )
\end{equation}
where $\mathcal{N}$ is the depth-sorted all Gaussian set, $\alpha'_i$ and $\mathbf{c}'_i$ are the opacity and color after deformation. 
In the hard stage, after hard routing result 
$\hat{l} $,
each Gaussian is exclusively assigned to one category. Let $S_l = \{\, i \mid \hat{l}_i = l \,\}$ denote the Gaussians assigned to category $l$. The category-specific image is then rendered from only $S_l$ using parameters predicted by branch $\mathcal{F}_l$:
\begin{equation}
I_l =
\sum_{i \in S_l}
\alpha'_i\,c'_i
\prod_{j < i,\, j \in S_l} ( 1 - \alpha'_j )
\end{equation}
where the product runs over the depth ordering within set $S_l$. 

The composite rendering $I_{\mathrm{total}} $ in either stage is obtained  over all updated Gaussians:
\begin{equation}
I_{\mathrm{total}} = \sum_{i \in \mathcal{N}}  \alpha_i' \,c_i' \prod_{j<i} \big(1 - \alpha_j'\big) 
\end{equation}

\subsection{Category-level Control }
\label{sec:control}
To ensure high‑quality rendering and decomposition in each category branch, we design distinct supervision strategies: brightness control, motion‑flow control, and mask control.

\textbf{Brightness control} keeps the background clean. Casual captures often suffer from lighting swings that blur geometry and cause shading artifacts. Although SH coefficients can model non‑Lambertian effects, they may misinterpret illumination changes as motion, breaking background consistency \cite{degauss}. We address this with a brightness‑aware mask in the background branch to absorb illumination changes and suppress motion ghosts. The raw mask is rasterized from Gaussian attribute $b$:
\begin{equation}
I_{\mathrm{B}} = \sum_{i \in \mathcal{N}}  \alpha_i' \,b_i' \prod_{j<i} \big(1 - \alpha_j'\big) 
\end{equation}
To handle extreme lighting, we apply a piecewise‑linear activation to obtain $\hat{I_{\mathrm{B}}}$:
\begin{equation}
\hat{I_{\mathrm{B}}} =
\begin{cases}
I_{\mathrm{B}}+0.5, & 0\leq{I_{\mathrm{B}}}\leq0.75, \\
k\left(I_{\mathrm{B}}-0.75\right)+1.25, & 0.75<I_{\mathrm{B}}\leq1 
\end{cases}
\end{equation}
where $k=35$ controls over‑brightness. The final background is $I_{bg} = \hat{I_{\mathrm{B}}}*I_{bg}$.

\textbf{Motion‑flow control} targets dynamic regions (hands/objects). We compute ground‑truth optical flow $\mathbf{F}_{t \rightarrow t+1}^{gt}$ from input frames and camera‑induced flow $\mathbf{F}_{t \rightarrow t+1}^{cam}$ from estimated pose. The dynamic flow is:
\begin{equation}
\mathbf{F}^m = \mathbf{F}_{t \rightarrow t+1}^{gt} - \mathbf{F}_{t \rightarrow t+1}^{cam}.
\end{equation}
Predicted flow $\hat{\mathbf{F}}^m$ from dynamic branches is supervised by:
\begin{equation}
\mathcal{L}_\mathrm{{flow}} = \lVert \hat{\mathbf{F}}^m - \mathbf{F}^m \rVert_1.
\end{equation}
This enforces accurate motion modeling in dynamic areas \cite{motiongs}.

\textbf{Mask control} enforces spatially‑aware supervision for all branches.  
Let $\mathbf{M}_{hand}$ and $\mathbf{M}_{obj}$ be binary masks, the mask‑weighted RGB and opacity losses are:
\begin{align}
\mathcal{L}_{\mathrm{rgb}}^{l} &= \big\| I_l \odot \mathbf{M}_l - I_{\mathrm{gt}} \odot \mathbf{M}_l \big\|_1, \\
\mathcal{L}_{\alpha}^{l} &= \big\| \alpha_l - \mathbf{M}_l \big\|_1
\end{align}
To avoid cross‑branch contamination, gradients are zeroed out in regions covered by other branches, using morphological expansion of their masks ($\mathbf{M}_{\mathrm{occ}}$):
\begin{equation}
\frac{\partial \mathcal{L}}{\partial I_l} \leftarrow \frac{\partial \mathcal{L}}{\partial I_l} \odot \big(1 - \mathrm{dilate}(\mathbf{M}_{\mathrm{occ}})\big),
\end{equation}

The overall loss is:
\begin{equation}
    \mathcal{L} = \mathcal{L}_1 + \mathcal{L}_{flow} + \sum_{l}(\mathcal{L}_{\mathrm{rgb}}^{l} + \mathcal{L}_{\alpha}^l + \mathcal{L}_\mathrm{SSIM}^{l}+ \mathcal{L}_\mathrm{entropy}^{l})
\end{equation}

\section{Experiment}
\label{sec:Experiment}

\subsection{\textbf{Experimental Settings}}
\textbf{Implementation Details} Our PyTorch-based implementation runs on a single RTX 3090 GPU. Scene boundaries and Gaussians are initialized from COLMAP \cite{colmap} point clouds, with \cite{egohos} and \cite{track} used for hand and object segmentation. Training comprises 10k soft iterations—starting with a 1k-iteration warm-up focusing only on probabilistic classification—and 10k hard iterations where each Gaussian is updated in its assigned deformation branch.

\textbf{Datasets} We take sequences from various Egocentric video datasets including HOI4D \cite{hoi4d}, Epic-Field \cite{epic} and Hot3D \cite{hot3d}.

\begin{figure}[thbp]
    \centering
    \includegraphics[width=1\linewidth]{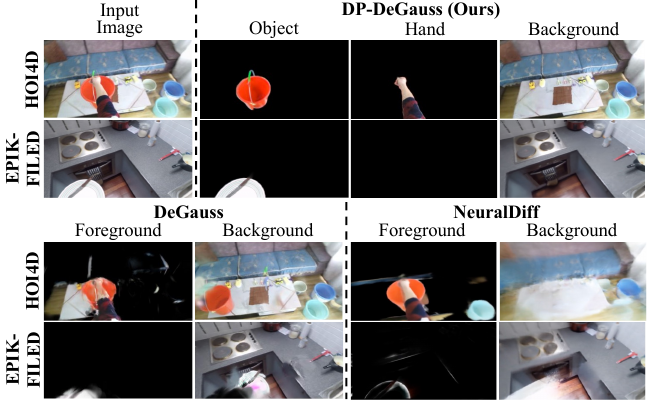}
  \caption{Visual comparison of scene decomposition into background, object, and hand components. Our method achieves clean, fine-grained decomposition with accurate boundaries and fewer artifacts.}
    \label{fig:Decomposition}
    \vspace{-3mm}
\end{figure}

\begin{table}[h]
\centering
\caption{Quantitative results of full reconstruction.}
\setlength{\tabcolsep}{4pt} 
\renewcommand{\arraystretch}{1.15} 
\resizebox{\linewidth}{!}{
\begin{tabular}{l|l|c c c c c}
\toprule
Dataset & Metric & \shortstack{4DGaussians\\ \cite{4dgs}} & \shortstack{MotionGS\\\cite{motiongs}} & \shortstack{NeuralDiff\\\cite{neuraldiff}}  & \shortstack{DeGauss\\\cite{degauss}} & \shortstack{\textbf{DP-DeGauss} \\\textbf{(Ours)}} \\
\midrule
\multirow{3}{*}{HOI4D}
& PSNR$\uparrow$& \underline{33.42} & 26.09 & 30.45& 31.52 & \textbf{33.69} \\
& SSIM$\uparrow$& 0.950 & 0.901 & 0.904& \underline{0.951} & \textbf{0.952} \\
& LPIPS$\downarrow$& \textbf{0.047} & 0.156 & 0.114& 0.042 & \underline{0.043} \\
\cmidrule(l){1-7}
\multirow{3}{*}{\shortstack{EPIC-\\FIELD}}
& PSNR$\uparrow$& 33.69& 27.83 & \underline{33.87}& 31.86 & \textbf{34.60} \\
& SSIM$\uparrow$& \underline{0.936}& 0.871 & 0.934& 0.929 & \textbf{0.941} \\
& LPIPS$\downarrow$ & 0.045& 0.152 & \textbf{0.043}& 0.066 & \underline{0.055} \\
\cmidrule(l){1-7}
\multirow{3}{*}{Hot3D}
& PSNR$\uparrow$& \underline{25.87} & 23.86 & 21.50& 25.72 & \textbf{26.12} \\
& SSIM$\uparrow$& 0.667 & 0.703& 0.700& \underline{0.704} & \textbf{0.711} \\
& LPIPS$\downarrow$& 0.282 & 0.316 & 0.301& \underline{0.237} & \textbf{0.262} \\
\bottomrule
\end{tabular}
}
\label{table:score}
\vspace{-3mm}
\end{table}
\subsection{Experiment Results}

We primarily present our results from the perspective of decoupling the background–object–hand components in terms of the overall scene reconstruction quality.

For \textbf{reconstruction quality}, we present qualitative comparisons with baseline methods in Fig.\ref{fig:Reconstruction}. Specifically, 4DGaussians\cite{4dgs} and MotionGS\cite{motiongs} only focus on full-scene reconstruction, while Neuraldiff\cite{neuraldiff} and DeGauss\cite{degauss} perform both reconstruction and decomposition. Our method preserves significantly more fine-grained details in the dynamic hand and object branches. Furthermore, for both dynamic branches and the static background, our approach effectively reduces motion blur artifacts and scene holes, clean in background and detailed in object and hand. 
For quantitative evaluation, we  select 3 sequences from each dataset and report the average results in Table\ref{table:score}, our method perform well in all metrics.

For \textbf{decomposition performance}, we compare with \cite{neuraldiff}, which separates foreground and background from egocentric videos, and \cite{degauss}, the most current  Gaussian-based reconstruction and decomposition method, as shown in Fig.\ref{fig:Decomposition} (The results of Hot3D dataset is already shown in Fig.\ref{fig:teaser}). Both baselines are limited to binary foreground–background separation, often misclassifying objects that are static in a single frame but dynamic over time, or even failing to separate forground and background. In addition, their background reconstructions are typically blurry, lacking fine details. They also struggle to distinguish hands from nearby dynamic objects, resulting in boundary leakage and inaccurate segmentation. In contrast, our method achieves fine-grained separation of hands, objects, and background, delivering cleaner decomposition and fewer artifacts.

\begin{figure}[t]
    \centering
    \includegraphics[width=1\linewidth]{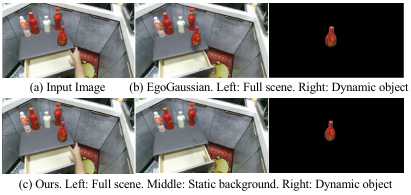}
    \caption{Visual comparison with EgoGaussian~\cite{egogaussian} on full-scene and object–background-separated reconstruction.}
    \label{fig:egogaussian}
    \vspace{-3mm}
\end{figure}

We additionally compare our method with EgoGaussian\cite{egogaussian}, which is designed for fine-grained modeling of object poses and trajectories. It represents one of the most recent and best-performing approaches for reconstructing both entire scene and object parts. However, it excludes hand when reconstructing the full scene, which does not fully align with our task, and its training time exceeds 24 hours, whereas our method requires only about 2 hours. We still present comparisons of full-scene reconstruction and object-background-separated reconstruction in Fig.\ref{fig:egogaussian}. Our method achieves comparable results, even with finer object details, demonstrating that it delivers both high efficiency and strong performance.


\subsection{Ablation Study}

\begin{figure}[h]
    \centering
    \includegraphics[width=1\linewidth]{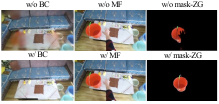}
    \caption{Ablation studies on brightness control, motion flow control and zero gradients on mask.}
    \label{fig:ablation}
\end{figure}

We conduct ablation studies on category-level controls in Fig.\ref{fig:ablation}.
For background decomposition, Brightness Control (BC) effectively removes non-background elements and ghosting artifacts left by dynamic objects.
Motion flow  (MF) helps reconstruct dynamic regions, such as the hand in the figure.
Applying Zero Gradients within masked regions (mask-ZG) during loss computation helps recover occluded parts of objects; otherwise, visible defects remain.
\vspace{-3mm}


\section{Conclusion}

We proposed DP-DeGauss, a dynamic probabilistic Gaussian decomposition framework from soft to hard for egocentric 4D reconstruction with explicit background–hand–object separation. By combining unified initialization, learnable category probabilities, and category-level controls, our method produces high-quality, fine-grained reconstructions and decomposition in challenging egocentric scenarios. In future, we will extend DP-DeGauss to diverse egocentric scenarios, improving adaptability to complex interactions.


\bibliographystyle{IEEEbib}
\bibliography{strings,refs}

\end{document}